\documentclass[10pt,twocolumn,letterpaper]{article}

\usepackage{cvpr}
\usepackage{times}
\usepackage{epsfig}
\usepackage{graphicx}
\usepackage{amsmath}
\usepackage{amssymb}
\usepackage{pgfplots}
\usepackage{tabu}
\usepackage{flushend}
\usepackage{authblk}
\usepgfplotslibrary{groupplots}

\renewcommand{\paragraph}[1]{\medskip\noindent\textbf{#1}\enskip}

\usepackage[pagebackref=true,breaklinks=true,letterpaper=true,colorlinks,bookmarks=false]{hyperref}

\cvprfinalcopy 

\def\cvprPaperID{3112} 

\ifcvprfinal\fi

\usepackage{tikz}
\usetikzlibrary{shapes,arrows,positioning,calc}
\usepackage{pgfplots}

\pgfplotsset{grid=major, legend style={font=\small}, y label style={at={(axis description cs:0.1,.5)},anchor=south}, enlargelimits=false, }
\usetikzlibrary{positioning}

\pgfplotsset{yticklabel style={text width=3em,align=right}}

\pgfplotscreateplotcyclelist{mycyclelistA}{
red,every mark/.append style={fill=red!80!black},mark=square*\\
red,densely dashed,every mark/.append style={solid,fill=red!80!black},mark=*\\
brown!60!black,every mark/.append style={fill=brown!80!black},mark=otimes*\\
blue,every mark/.append style={fill=blue!80!black},mark=*\\
blue,every mark/.append style={fill=blue!80!black},mark=diamond*\\
blue,densely dashed,mark=star,every mark/.append style=solid\\
brown!60!black,densely dashed,every mark/.append style={
solid,fill=brown!80!black},mark=square*\\
black,mark=star\\
black,densely dashed,every mark/.append style={solid,fill=gray},mark=otimes*\\
red,densely dashed,every mark/.append style={solid,fill=red!80!black},mark=diamond*\\
}

\pgfplotscreateplotcyclelist{mycyclelistB}{
red,every mark/.append style={fill=red!80!black},mark=square*\\
red,densely dashed,every mark/.append style={solid,fill=red!80!black},mark=*\\
blue,every mark/.append style={fill=blue!80!black},mark=*\\
blue,every mark/.append style={fill=blue!80!black},mark=triangle*\\
blue,every mark/.append style={fill=blue!80!black},mark=diamond*\\
blue,densely dashed,mark=star,every mark/.append style=solid\\
brown!60!black,every mark/.append style={fill=brown!80!black},mark=otimes*\\
brown!60!black,densely dashed,every mark/.append style={
solid,fill=brown!80!black},mark=square*\\
black,densely dashed,every mark/.append style={solid,fill=gray},mark=otimes*\\
red,densely dashed,every mark/.append style={solid,fill=red!80!black},mark=diamond*\\
}

\pgfplotscreateplotcyclelist{mycyclelistC}{
red,every mark/.append style={fill=red!80!black},mark=square*\\
blue,every mark/.append style={fill=blue!80!black},mark=*\\
blue,densely dashed,mark=star,every mark/.append style=solid\\
}

\usepackage{soul}
\usepackage{color}

\usepackage{stmaryrd}
\usepackage{booktabs}
\usepackage{paralist}
\usepackage{graphicx}
\usepackage{amssymb}
\usepackage{amsmath}
\usepackage{amsthm}
\usepackage{graphicx}
\usepackage{bm}

\def\R{\mathbb{R}}
\def\ms{\mathcal}
\newcommand{\col}[1]{\textsc{\textrm{col}} \left ( #1 \right )}
\newcommand{\Tr}[0]{{\ensuremath{\textrm{T}}}}
\renewcommand{\paragraph}[1]{\medskip\noindent\textbf{#1}\enskip}

\newcommand{\bmt}[1]{\tilde{\bm{#1}}} 

\def\subic{{\sc SuBiC}}
\def\subicI{{\subic-I}}
\def\subicR{{\subic-R}}
\def\subicJ{{\subic-J}}
\def\IVFPQ{{IVF-PQ}}
\def\IMIPQ{{IMI-PQ}}

\def\oxf{{Oxford5K}}
\def\paris{{Paris6K}}
\def\holidays{{Holidays}}
\def\flickr{{Flickr100K}}

\def\landm{{Landmarks}}
\def\oxff{{Oxford105K}}
\def\parisf{{Paris106K}}
\def\voc{{Pascal VOC}}
\def\caltech{{Caltech-101}}
\def\relu{{{ReLU}}}

\DeclareMathOperator*{\argmin}{argmin}
\DeclareMathOperator*{\argmax}{argmax}

\newcommand{\expp}[1]{\exp \left ( #1 \right )}
\def\E{\textrm{E}}
\newcommand{\ellT}[1]{\ell \left ( #1 \right )}

\newcommand{\ellEs}[1]{\ell_{\mathrm{E}}(#1)}
\newcommand{\ellB}[1]{\ell_{\mathrm{B}} \left ( #1 \right )}
\newcommand{\ellSx}[3]{\ell_{\mathrm{S}}^{#1,#2} \big( #3 \big)}
\newcommand{\ellS}[1]{\ellSx{\mu}{\gamma}{#1}}

\newcommand{\figref}[1]{Fig. \ref{#1}}
\newcommand{\sxnref}[1]{\S\ref{#1}}

\definecolor{myblue}{HTML}{85C1E9}
\definecolor{myyellow}{HTML}{F7DC6F}
\definecolor{mygreen}{HTML}{7DCEA0}
\definecolor{fitcolor}{HTML}{EAECEE}
\usetikzlibrary{positioning,arrows,fit,calc,shadows}
\tikzset{
  diagonal fill/.style 2 args={fill=#2, path picture={
      \fill[#1, sharp corners] (path picture bounding box.south west) -|
      (path picture bounding box.north east) -- cycle;}},
  reversed diagonal fill/.style 2 args={fill=#2, path picture={
      \draw[gray, fill=#1, sharp corners] (path picture bounding box.north west) |-
      (path picture bounding box.south east) -- cycle;}}}
\tikzset{%
  train/.style={fill=myblue},
  test/.style={fill=myyellow},
  traintest/.style={fill=mygreen},
  fit node/.style={fill=fitcolor, drop shadow},
  block/.style={draw, fill=blue!20!white, rounded corners, rectangle,
    minimum height=2em, minimum width=2em, font=\footnotesize\scshape, align=center, drop shadow},
  W/.style={draw, rectangle, minimum height=1.5em, minimum width=1.5em, drop shadow, traintest},
  bbox/.style={draw=gray, rounded corners, rectangle},
  activ/.style={font=\tiny\itshape,at start,inner sep=2pt},
  fit label/.style={inner sep=.3em,font=\small\itshape},
  input/.style={inner sep=0pt},
  output/.style={inner sep=0pt},
  sum/.style = {draw, fill=white, circle, minimum size=.8em, node distance=5em, inner sep=0pt,traintest},
  pinstyle/.style = {pin edge={to-,thin,black}}
}

\def\mv{\bm}

\begin{document}

\title{Learning a Complete Image Indexing Pipeline}

\author[1,2]{Himalaya Jain}
\author[3]{Joaquin Zepeda}
\author[1]{Patrick P\'erez}
\author[2]{R\'emi Gribonval}
\affil[]{Technicolor, Rennes, France $\quad$ $^2$INRIA Rennes, France \quad $^3$Amazon, Seattle, WA}


\maketitle

\begin{abstract}
To work at scale, a complete image indexing system comprises two components: An inverted file index to restrict the actual search to only a subset that should contain most of the items relevant to the query; An approximate distance computation mechanism to rapidly scan these lists. While supervised deep learning has recently enabled improvements to the latter, the former continues to be based on unsupervised clustering in the literature. In this work, we propose a first system that learns both components within a unifying neural framework of structured binary encoding.
\end{abstract}
\section{Introduction}

Decades of research have produced powerful means to extract features from images, effectively casting the visual comparison problem into one of distance computations in abstract spaces. Whether engineered
or trained using convolutional deep networks, 
such vector representations are at the core of all content-based visual search engines. This applies particularly to example-based image retrieval systems where a query image is used to scan a database for images that are similar to the query in some way:
in that they are the same image but one has been edited (\emph{near duplicate detection}), or because they are images of the same object or scene (\emph{instance retrieval}), or because they depict objects or scenes from the same semantic class (\emph{category retrieval}).

Deploying such a visual search system requires conducting nearest neighbour search in a high-dimensional feature space. Both the  dimension of this space and the size of the database can be very large, which imposes severe constraints if the system is to be practical in terms of storage (memory footprint of database items) and of computation (search complexity). Exhaustive exact search must be replaced by \textit{approximate, non-exhaustive} search.
To this end, two main complementary methods have emerged, both relying on variants of unsupervised vector quantization (VQ).
The first such method, introduced by Sivic and Zisserman \cite{sivic2003video} 
is the inverted file system. Inverted files rely on a partitioning of the feature space into a set of mutually exclusive bins. Searching in a database thus amounts to first assigning the query image to one or several such bins, and then ranking the resulting shortlist of images associated to these bins using the Euclidean distance (or some other distance or similarity measure) in feature space.

The second method, introduced by Jegou \etal~\cite{jegou2011product}, consists of using efficient approximate distance computations as part of the ranking process. This is enabled by feature encoders producing compact representations of the feature vectors that further do not need to be decompressed when computing the approximate distances. This type of approaches, which can be seen as employing block-structured binary representations, superseded the (unstructured) binary hashing schemes 
that dominated approximate search.

%
%
Despite its impressive impact on the design of image representations \cite{gordo2016deep,arandjelovic2016netvlad,Gong2014,Sharif}, supervised deep learning is still limited in what concerns the approximate search system itself.
Most recent efforts focus on supervised deep binary hashing schemes, as discussed in the next section. 
As an exception, the work of Jain \etal~\cite{jain2017subic} employs a block-structured approach inspired by the successful compact encoders referenced above. 
Yet the binning mechanisms that enable the usage of inverted files, and hence large-scale search, have so far been neglected.

In this work we introduce a novel supervised inverted file system along with a supervised, block-structured encoder that together specify a complete, supervised, image indexing pipeline. Our design is inspired by the two methods of successful indexing pipelines described above, while borrowing ideas from \cite{jain2017subic} to implement this philosophy.
%

Our main contributions are as follows: (1) We propose the first, to our knowledge, image indexing system to reap the benefits of deep learning for both data partitioning and feature encoding; (2) Our data partitioning scheme, in particular, is the first to 
replace unsupervised VQ by a supervised approach; (3) We take steps towards learning the feature encoder and inverted file binning mechanism simultaneously as part of the same learning objective; (4) We establish a wide margin of improvement over the existing baselines employing state-of-the art deep features, feature encoders and binning mechanism.
\section{Background}
\label{sec:background}



\paragraph{Approximating distances through compact encoding}
Concerning approximate distance computations, two main approaches exist. Hashing methods \cite{wang2014hashing}, on the one hand, 
employ Hamming distances between binary hash codes.
Originally unsupervised, these methods have recently benefited from progress in deep learning \cite{xia2014supervised,zhang2015bit,zhao2015deep,lai2015simultaneous,lin2015deep,liu2016deep,do2016learning}, leading to better systems for category retrieval in particular.
Structured variants of VQ, on the other hand, produce fine-grain approximations of the high-dimensional features themselves through very compact codes \cite{babenko2014additive,ge2013optimized,ge2014product,jain2016approximate,jegou2011product,kalantidis2014locally,nourouzi2013cartesian,zhang2014composite} that enable look-up table-based efficient distance computations. 
Contrary to recent hashing methods, VQ-based approaches have not benefited from supervision so far. However, Jain \etal \cite{jain2017subic} recently proposed a supervised deep learning approach that leverages the advantages of structured compact encoding and yields state-of-the-art results on several retrieval tasks. Our work extends this supervised approach towards a complete indexing pipeline, that is, a system that also includes an inverted file index.    
 


\paragraph{Scanning shorter lists with inverted indexes} For further efficiency, approximate search is further restricted to a well chosen fraction of the database. This pruning is carried out by means of an Inverted File (IVF), which relies on a partitioning of the feature space into Voronoi cells defined using $K$-means clustering \cite{jegou2008hamming,babenko12inverted}. Two things should be noted: The method to build the inverted index is unsupervised and it is independent from the way subsequent distance approximations are conducted (\eg, while VQ is used to build the index, short lists can be scanned using binary embeddings \cite{jegou2008hamming}). In this work, we propose a unifying supervised framework. Both the inverted index and the encoding of features are designed and trained together for improved performance. In the next section, we expose in more detail the existing tools to design IVF/approximate search pipelines, before moving to our proposal in Section \ref{sec:approach}.

\section{Review of image indexing} \label{sec:review}

Image indexing systems are based on two main components: \text{(i)} an \emph{inverted file} and \textit{(ii)} a \emph{feature encoder}. In this section we describe how these two main components are used in image indexing systems, thus laying out the motivation for the method we introduce in \sxnref{sec:approach}.




\paragraph{Inverted File (IVF)} An inverted file relies on a partition of the database into mutually exclusive bins, a subset of which is searched at query time. The partitioning is implemented by means of VQ \cite{sivic2003video,jegou2011product,babenko12inverted}:  Given a vector $\bm x \in \R^d$ and a codebook $\bm D = [\bm d_k \in \R^d]_{k=1}^N$, the VQ representation of $\bm x$ in $\bm D$ is obtained by solving\footnote{Notation: We denote $[ \bm v_1 , \ldots,  \bm v_K] = [ \bm v_k \in \R^d]_{k=1}^M$ the matrix in $\R^{d \times M}$ having columns $\bm v_k \in \R^d$, or simply $[\bm v_k]_k$. For scalars $a_k$, $[a_k]_k$ denotes a column-vector with entries $a_k$. The column vector obtained by stacking vertically vectors $\bm v_k$  is noted $\col{\bm v_1, \ldots, \bm v_K}$. We further let $\bm v[k]$ denote the $k$-th entry of vector $\bm v$.}
\begin{equation}\label{eq:bin assignment}
  n=\argmin\nolimits_k \| \bm x - \bm d_k \|_2^2, 
\end{equation}
where $n$ is  the \emph{codeword index} for $\bm x$ and $\bm d_n$ its \emph{reconstruction}. Given a database $\{ \bm x_i \}_i$ of image features, and letting $n_i$ represent the codeword index of $\bm x_i$, the database is partitioned into $N$ index bins
$\ms B_n$.
These bins, stored along with metadata that may include the features $\bm x_i$ or a compact representation thereof, is known as an inverted file. At query time, 
the bins are ranked by decreasing pertinence $n_1, \ldots, n_N$ 
relative to the query feature $\bm x^*$ so that 
\begin{equation} \label{eq:dist sorting}
  \| \bm x^* - \bm d_{n_1} \| \leq \ldots \leq \| \bm x^* - \bm d_{n_{N}} \|,
\end{equation}
\ie, by increasing order of reconstruction error. Using this sorting, one can specify a target number of images $T$ to retrieve from the database and search only the first $B$ bins so that $\sum_{k=1}^{B - 1} \left | \ms B_{n_k} \right | \leq T \leq \sum_{k=1}^{B} \left | \ms B_{n_k} \right |$.

It is important to note that all existing state-of-the-art indexing methods 
employ a variant of the above described mechanism that relies on $K$-means-learned codebooks $\bm D$. To the best of our knowledge, ours is the first method to reap the benefits of deep learning to build an inverted file.

\paragraph{Feature encoder} The inverted file 
outputs a shortlist of images with indices in $\bigcup_{k=1}^B\ms B_{n_k}$, which needs to be efficiently ranked 
in terms of distance to the query 
This is enabled by compact feature encoders that allow rapid distance computations without decompressing features. 
It is important to note that the storage bitrate of the encoding affects -- besides storage cost -- search speed, as higher bitrates means that bins need to be stored in secondary storage, where look-up speeds are a significant burden.

State-of-the art image indexing systems use feature encoders that employ a residual approach: A residual is computed from each database feature $\mv x$ and its reconstruction $\bm d_n$ obtained as part of the inverted file bin selection in \eqref{eq:bin assignment}:
\begin{equation} \label{eq:db residual}
  \bm r_n = \bm x - \bm d_n.
\end{equation}
This residual is then encoded using a very high resolution quantizer. Several schemes exist \cite{chen2010approximate,jegou2011product} that exploit structured quantizers to enable low-complexity, high-resolution quantization, and herein we describe \emph{product quantizers} and related variants \cite{jegou2011product,nourouzi2013cartesian,ge2013optimized}. Such vector quantizers employ a codebook  $\mv C \in \R^{d \times K^M}$ with codewords that are themselves additions of codewords from $M$ smaller \emph{constituent} codebooks $\bm C_m = \left [ \bm c_{m,k} \right ]_k \in \R^{d \times K}, m=1,\ldots,M$,  that are orthogonal ($\forall m \neq l, \bm C_m^\Tr \bm C_l = \bm 0$):
\begin{equation} \label{eq:constituent}
  \bm C = \Big[ \sum\nolimits_{m=1}^M \bm c_{m,k_m} \Big]_{(k_1, \ldots, k_M) \in (1,\ldots,K)^M}.
\end{equation}
Accordingly, an encoding of $\bm r$ in this structured codebook is specified by the indices $(k_1, \ldots, k_M)$ which uniquely define the codeword $\bm c$ from $\bm C$, \ie, the reconstruction of $\bm r$ in $\bm C$. Note that the bitrate of this encoding is $M \log_2(K)$.

\paragraph{Asymmetric distance computation} Armed with such a representation for all database vectors, one can very efficiently compute an approximate distance between a query $\bm x^*$ and all database features $\bm x \in \{ \bm x_i, i \in \cup_{k=1}^{B} \ms B_{n_k} \}$ in top-ranked bins. 
The residual of $\bm x^*$ for bin $\ms B_n$ is 
\begin{equation} \label{eq:query residual}
  \mv r_n^* = \bm x - \bm d_n
\end{equation}
and the approach is asymmetrical in that this uncompressed residual is compared to the compressed, reconstructed residual representation $\bm c$  of the database vectors $\bm x$ in bin $\ms B_n$ using the distance
\begin{equation}\label{eq:approx}
 \| \bm r_n^* -  \bm c \|_2^2 = \sum_{m=1}^M \| \bm r_n^* - \bm c_{m,k_m} \|_2^2.    
\end{equation}
We define the look-up tables (LUT)
\begin{equation} \label{eq:zluts}
  \bm z_{n,m} \triangleq \left [ \; \| \mv r_n^* - \mv c_{m,k} \|_2^2 \; \right ]_k \in \R^K 
\end{equation}
containing the distances between $\mv r_n^*$ and all codewords of $\bm C_m$. Building these LUTs enables us to compute \eqref{eq:approx} using $\sum_{m=1}^M \bm z_m[k_m]$, 
an operation that requires only $M$ table look-ups and additions, establishing the functional benefit of the encoding $(k_1,\ldots,k_M)$.

To gain some insight into the above encoding, 
consider the one-hot representation $\bm b_m$ of the indices $k_m$ given by
\begin{equation}
  \bm b_m = \big [ \llbracket l = k_m \rrbracket \big ]_l \in \ms K_K,
\end{equation}
where $\llbracket \cdot \rrbracket$ denotes the Iverson brackets and
\begin{equation}
  \ms K_K \triangleq \{ \bm a \in \{0,1\}^K , \|\bm a\|_1 = 1\}.
\end{equation}

Using stacked column vectors
\begin{align}
  \bm b &=  \col{\bm b_1, \ldots, \bm b_M}  \in \ms K_K^M \textrm{ and } \label{eq:block code} \\
  \bm z_n &= \col{\bm z_{n,1}, \ldots, \bm z_{n,M}}  \in \R_+^{MK}, \label{eq:query mapping}
\end{align}
distance \eqref{eq:approx} can be expressed as follows:
\begin{equation} \label{eq:lut}
  \| \bm r_n^* -  \bm c \|_2^2  = \bm z_n^\Tr \bm b.
\end{equation}
Namely, computing approximate distances between a query $\bm x^*$ and the database features $\bm x \in \{ \bm x_i, i\in \ms B_n\}$ amounts to computing an inner-product between a bin-dependent mapping $\bm z_n \in \R^{MK}$ of the query feature $\bm x^*$ and a block-structured binary code $\bm b \in \ms K_M^K$ derived from $\bm x$.
A search then consists of computing all such approximate distances for the $B$ most pertinent bins
and then sorting the corresponding images in increasing order of these distances.

It is worth noting that most of the recent supervised binary encoding methods 
\cite{xia2014supervised,zhang2015bit,zhao2015deep,lai2015simultaneous,lin2015deep,liu2016deep,do2016learning} do not use structured binary codes of the form $\bm b$ in \eqref{eq:lut}. The main exception being \subic\ \cite{jain2017subic}, which further uses a sorting score that is an inner product of the same form as  \eqref{eq:lut}.


\section{A complete indexing pipeline} \label{sec:approach}
The previous section established how state-of-the-art large-scale image search systems rely on two main components: an inverted file
and a functional residual encoder that produces block-structured binary codes.
While compact binary encoders based on deep learning have been explored in the literature, inverted file systems continue to rely on unsupervised $K$-means codebooks.

In this section we first revisit the \subic\ encoder \cite{jain2017subic}, and then show how it can be used to implement a complete image indexing system that employs deep learning methodology both at the IVF stage and compact encoder stage.

\subsection{Block-structured codes}
\begin{figure}[t]
  \centering
  \begin{tikzpicture}[auto, node distance=4em and 4em, on grid, >=latex', font=\small]

  \node[block, traintest] (baseCNN) {Base\\CNN};
  
  \node[W, right=1.5em of baseCNN.east] (W) {$\bm W$};
  \node[block, train, right=of W.east](block softmax){Block\\SoftMax};
  \node[block, test, above=1em of block softmax.north](block one hot){Block\\One-Hot};

  \draw [->] (W)  -- node[activ,above right]{ReLU} node[below]{$\bm z \in \R^{KM}_+$} (block softmax);
  \draw [->] ($(block softmax.west)-(1em,0)$) |- (block one hot);
  \draw [->] (block softmax.east) --  node[above right]{$\bmt{b} \in \Delta_{K}^M$}   +(2em,0) ;
  \draw [->] (block one hot.east) -- node[above right]{$\bm b \in \ms K_{K}^M$} +(2em,0);

  \draw [->] ($(baseCNN.west) - (1.5em,0)$) -- node[above]{$\bm I$} (baseCNN);
  \draw [->] (baseCNN.east) -- node[above]{$\bm x$} (W.west);
  
\end{tikzpicture}

  \caption{The \subic\ encoder operates on the feature vector $\bm x$ produced by a CNN to enable learning of (relaxed) block-structured codes ($\bmt b$) $\bm b$. Blue, yellow, and green blocks are active, respectively, only at training time, only at testing time and at training/testing times. }
  \label{fig:subic}
\end{figure}
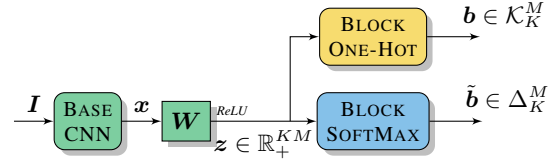
\begin{figure}[t]
  \centering
  \includegraphics[height=0.4\columnwidth]{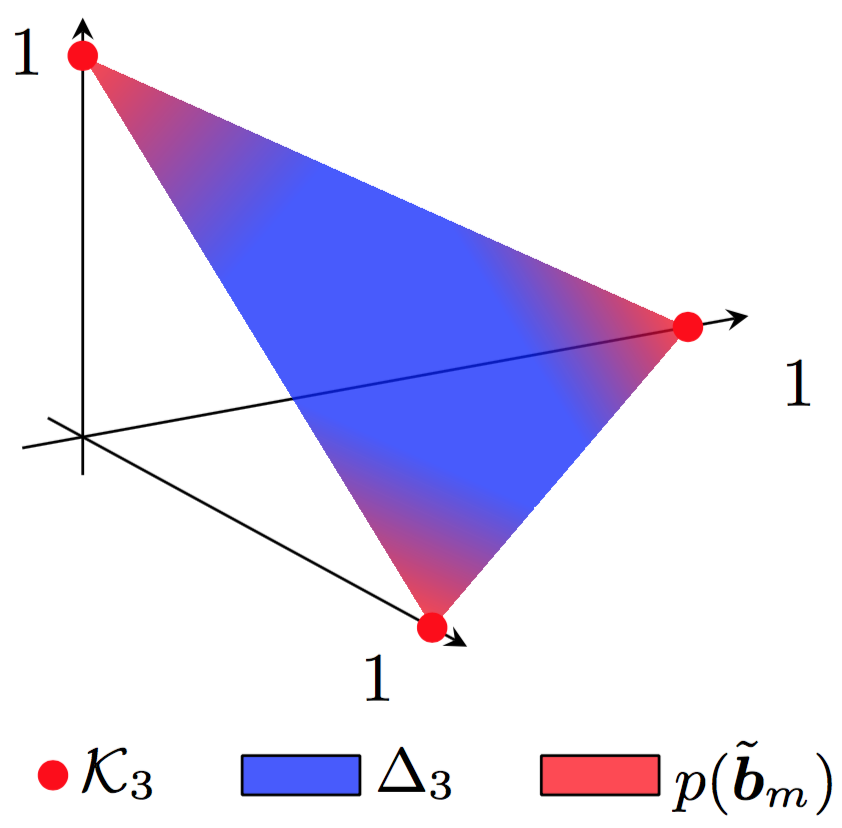} 
  \caption{ The discrete set $\ms K_3$ of one-hot encoded vectors, its convex-hull $\Delta_3$, and the distribution of relaxed blocks $\bmt b_m$ enforced by the \subic\ entropy losses. Omitting the negative batch entropy loss \eqref{eq:ngtv batch entropy loss} would result in situations where $p (\bmt b_m)$ is concentrated near only $k<3$ of the elements in $\ms K_3$.}
  \label{fig:K3D3}
\end{figure}
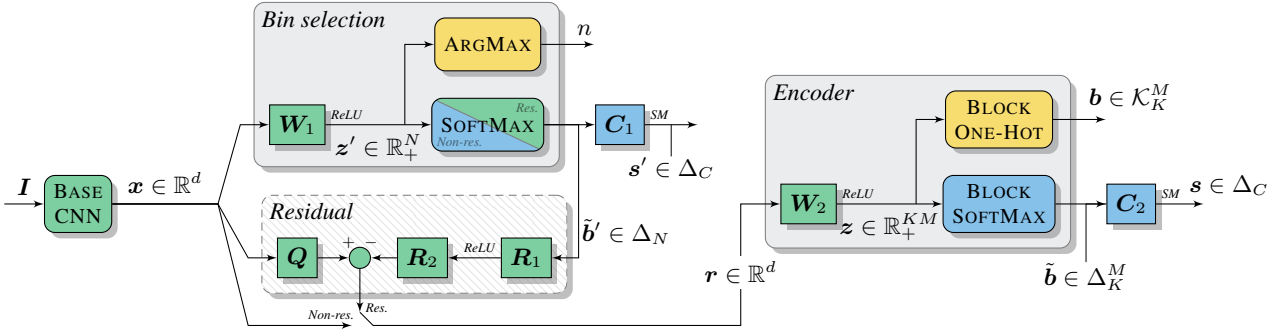
\begin{figure*}[!t]
  \centering
\usetikzlibrary{backgrounds,patterns}

\begin{tikzpicture}[auto, node distance=4em and 4em, on grid, >=latex', font=\small]
  
  \node[block, traintest] (baseCNN) {Base\\CNN};
  
  \node[W] (W1) at ($(baseCNN.east)+(7em,3em)$) {$\bm W_1$};
  \node[block, traintest, right=of W1.east, reversed diagonal fill={myblue}{mygreen}](softmax){SoftMax};
  \tikzset{small label/.style={font=\tiny\itshape,inner sep=2pt}}
  \node[anchor=north east, small label, text=mygreen!50!black] at (softmax.north east) {Res.};
  \node[anchor=south west, small label, text=myblue!50!black] at (softmax.south west) {Non-res.};
  \node[block, traintest, test, above=1em of softmax.north](argmax){ArgMax};
  \node[W, train, right=5em of softmax] (C1) {$\bm C_1$};
  \begin{scope}[on background layer]
    \node[draw,fit node,inner sep=2mm,bbox,fit=(W1) (softmax) (argmax)](bin sel) {};
  \end{scope}
  \node[anchor=north west, fit label] at (bin sel.north west) {Bin selection};
  
  \node[W, below=3.5em of W1.south](Q){$\bm Q$};
  \node[W, right=3em of Q.east](R2){$\bm R_2$};
  \node[sum](minus) at ($(Q)!0.5!(R2)$) {$$};
  \node[W, right=2em of R2.east](R1){$\bm R_1$};
  \coordinate[](Q top padding) at ($(Q)+(0,1.8em)$){};
  \begin{scope}[on background layer]
    \node[draw,fit node, preaction={fill,white,draw}, densely dashed, pattern=north west lines, pattern color=black!10!white, inner sep=2mm,bbox, fit=(Q) (R1) (Q top padding) ](residual) {};
  \end{scope}
  \node[anchor=north west, fit label] at (residual.north west) {Residual};

  \coordinate[](switch) at ($(minus |- residual.south) - (0,1.2em)$){};
  
  \node[W] at ($(softmax.east)+(10em,-3em)$) (W2) {$\bm W_2$};
  \node[block, train, right=of W2.east](block softmax){Block\\SoftMax};
  \node[W, train, right=5em of block softmax](C2){$\bm C_2$};
  \node[block, test, above=1em of block softmax.north](block one hot){Block\\One-Hot};
  \begin{scope}[on background layer]
    \node[draw,fit node,inner sep=2mm,bbox, fit=(W2) (block softmax) (block one hot)](encoder) {};
  \end{scope}
  \node[anchor=north west, fit label] at (encoder.north west) {Encoder};
  
  \def\spread{4em}
  \draw [->] ($(baseCNN.west) - (1.5em,0)$) -- node[above]{$\bm I$} (baseCNN);
  \draw [->] (baseCNN.east) -- node[above]{$\bm x \in \R^d$} +(\spread,0) -- ($(bin sel.west |- W1) -(.5em,0)$) -- (W1.west);
  \draw [->] (baseCNN.east) --                      +(\spread,0) -- ($(residual.west |- Q.west)-(.5em,0)$) -- (Q.west);

  \draw [->] (baseCNN.east) --                      +(\spread,0) -- ($(switch -| residual.west)-(.5em,0)$) -- ($(switch) - (0.3em,0)$);
  \draw [->] (minus.south) -- ($(switch) + (0,0.5em)$) node[small label,right,pos=0.9]{Res.} coordinate(residual on){};
  \draw [->] (residual on) -- ($(switch) + (0.5em,0)$) -| node[pos=0.72, anchor=center, fill=white, inner sep=2pt]{$\bm r \in \R^d$} ($(W2.west)-(1.5em,0)$) -- (W2);
  \node[small label, above left] at (switch) {Non-res.};
  
  \draw [->] (W1)  -- node[activ,above right]{ReLU} node[below]{$\bm z' \in \R^{N}_+$} (softmax);
  \draw [->] ($(softmax.west)-(1em,0)$) |- (argmax);
  \draw [->] (argmax.east) -- node[above right]{$n$} +(2em,0);
  \draw [->] (softmax.east) -- +(1.3em,0) |- node[pos=0.4,anchor=west,inner sep=1pt]{$\bmt{b}' \in \Delta_{N}$}    (R1.east) ;
  \draw [->] (softmax.east) -- (C1.west);
  \draw [->] (C1.east) -- node[activ,above right]  {SM} +(2em,0) node[below, inner sep=0pt, yshift=-1.2em, pos=0.5](sp){$\bm s' \in \Delta_C$};
  \draw [-,very thin] (sp.north) -- (C1.east -| sp);

  \draw [->] (Q) -- node[at end,above,font=\tiny]{$+$}(minus);  
  \draw [->] (R1) -- node[activ,above left]{ReLU} (R2);
  \draw [->] (R2) -- node[at end,above,font=\tiny]{$-$} (minus);

  \draw [->] (W2)  -- node[activ,above right]{ReLU} node[below]{$\bm z \in \R^{KM}_+$} (block softmax);
  \draw [->] ($(block softmax.west)-(1em,0)$) |- (block one hot);
  \draw [->] (block softmax.east) --  node[below,yshift=-2.2em,pos=0.6,inner sep=0](bt){$\bmt{b} \in \Delta_{K}^M$}   (C2);
  \draw [-, very thin] (C2.west) -| (bt.north);
  \draw [->] (C2.east) --node[activ,above right]{SM} node[above right]{$\bm s \in \Delta_{C}$} +(1.8em,0);
  \draw [->] (block one hot.east) -- node[above right]{$\bm b \in \ms K_{K}^M$} +(2em,0);

\end{tikzpicture}

  \caption{\textbf{Proposed indexing architecture.} The proposed indexing architecture consists of a bin selection component, a residual computation component, and a feature encoder. We use blocks with square corners (labeled with a weights matrix) to denote fully-connected linear operations, potentially followed by a \relu\ or softmax (SM) nonlinearity. Blue, yellow, and green blocks are active, respectively, only at training time, only at testing (\ie database indexing / querying) time and at training/testing times. The residual block can be disabled to define a new architecture, as illustrated by the switch at the bottom of the diagram. }
  \label{fig:block diagram}
\end{figure*}

The \subic\ encoder 
in Fig.\ \ref{fig:subic} was the first to leverage supervised deep learning to produce a block-structured code of the form $\bm b \in \ms K_K^M$ in \eqref{eq:block code}. At learning time, the method relaxes the block-structured constraint. Letting 
\begin{equation}
  \Delta_K = \big \{ \bm a \in \R_+^{K} \textrm{ s.t. } \sum\nolimits_{k} \bm a[k] = 1 \big \} 
\end{equation}
denote the convex hull of $\ms K_K$, said relaxation 
\begin{equation}
  \bmt b \in \Delta_K^M
\end{equation}
is enforced by means of a fully-connected layer of output size $KM$ and \relu\ activation with output $\bm z$ that is fed to a \emph{block softmax} non-linearity that operates as follows: Let $\bm z_m$ denote the $m$-th block of $\bm z  \in \R^{KM}$ such that $\bm z = \col{ \bm z_1, \ldots, \bm z_M}$. Likewise, let $\bmt b_m \in \Delta_K$ denote the $m$-th block of the relaxed code $\bmt b \in \Delta_K^M$. The \emph{block softmax} non-linearity operates by applying a standard softmax non-linearity to each block $\bm z_m$ of $\bm z$ to produce the corresponding block $\bmt b_m$ of $\bmt b$:
\begin{equation}\label{eqn:block softmax}
  \bmt b_m = \left [ \frac{\expp{\bm z_m[k]} }{ \sum_l \expp{\bm z_m[l]} } \right ]_k.
\end{equation}
At test time, the block-softmax non-linearity is replaced by a \emph{block one-hot} encoder that projects $\bmt b$ unto $\Delta_K^M$. In practice, this can be accomplished by means of one-hot encoding of the index of the maximum entry of $\bm z_m$:
\begin{equation}
  \bm b_m = \big [ \; \llbracket k = \argmax( \bm z_m) \rrbracket \; \big ]_k.
\end{equation}

The approach of \cite{jain2017subic} introduced two losses based on entropy that enforce the proximity of $\bmt b$ to $\ms K_K^M$. The entropy of a vector $\bm p \in \Delta_K$, defined as
\begin{equation}
  \E (\bm p) = \sum\nolimits_{k=1}^K \bm p[k] \log_{2} \left ( \bm p[k] \right ),
\end{equation}
has a minimum equal to zero for deterministic distributions $\bm p \in \ms K_K$, motivating the use of the \emph{entropy loss}
\begin{equation} \label{eq:entropy loss}
  \ellEs{\bmt b} \triangleq \sum\nolimits_{m=1}^M \E ( \bmt b_m)
\end{equation}
to enforce the proximity of the relaxed blocks $\bmt b_m$ to $\ms K_K$.
This loss on its own, however, could lead to situations where only some elements of $\ms K_K$ are favored (\cf \figref{fig:K3D3}), meaning that only a subset of the support of the $\bm b_m$ is used. 

Yet entropy likewise has a maximum of $\log_2(K)$ for uniform distributions $\bm p = \frac{1}{K} \bm 1$. This property can be used to encourage uniformity in the selection of elements of $\ms K_K$ by means of the \emph{negative batch entropy loss}, computed for a batch $\ms A = \{ \bmt b^{(i)} \}_i$ of size $|\ms A|$ using
\begin{equation}\label{eq:ngtv batch entropy loss}
  \ellB{\ms A} \triangleq -\sum\nolimits_{m=1}^M \E \Big( \frac{1}{| \ms A |} \sum_i \bmt b_m^{(i)}\Big).
\end{equation}
For convenience, we define the \subic\ loss computed on a batch $\ms A$ as the weighted combination of the two entropy losses, parametrized by the hyper-parameters $\mu,\gamma \in \R_+$:
\begin{equation}
  \ellS{\ms A} \triangleq \frac{\mu}{|\ms A|} \sum\nolimits_{\bmt b \in \ms A} \ellEs{\bmt b}  + \gamma \ellB{\ms A}.
  \label{eq:subic_loss}
\end{equation}

It is important to point out that, unlike the residual encoder  described in \sxnref{sec:review}, the \subic\ approach operates on the feature vector $\bm x$ directly. Indeed, the \subic\ method is only a feature encoder, and does not implement an entire indexing framework.

\subsection{A novel indexing pipeline}
We now introduce our proposed network architecture that uses the method of \cite{jain2017subic} described above to build an entire image indexing system. The system we design implements the main ideas of the state-of-the-art pipeline described in \sxnref{sec:review}.

Our proposed network architecture is illustrated in \figref{fig:block diagram}. The input to the network is the feature vector $\mv x$ consisting of activation coefficients obtained by running a given image $\bm I$ through a CNN feature extractor. We refer to this feature extractor as the \emph{base CNN} of our system.

Similarly to the design philosophy described in \S\ref{sec:review}, our indexing system employs an IVF and a residual feature encoder. Accordingly, the 
architecture in \figref{fig:block diagram} consists of two main blocks, \emph{Bin selection} and \emph{Encoder}, along with a \emph{Residual} block that links these two main components.

\paragraph{Bin selection} The first block, labeled \emph{Bin selection} in \figref{fig:subic} can be seen as a \subic\ encoder employing a single block (\ie $M=1$) of size $N$, with the block one-hot encoder substituted by an argmax operation. The block consists of a single fully-connected layer with weight matrix $\bm W_1$ and \relu\ activation followed by a second activation using softmax. When indexing a database image $\bm I$, this block is responsible for choosing the bin $\ms B_n$ that $\bm I$ is assigned to, using the argmax of the coefficients $\bm z'$.

Given a query image $\bm I^*$, the same binning block is responsible for sorting the bins 
in decreasing order of pertinence $\ms B_{n_1} \cdots \ms B_{n_N}$ 
using the coefficients $\bm z'^* \in \R^N_+$ so that
\begin{equation}\label{eq:sim sorting}
  \bm z'^*[n_1] \geq \ldots \geq \bm z'^*[n_N],
\end{equation}
in a manner analogous to \eqref{eq:dist sorting}.

\paragraph{(Residual) feature encoding}
Inspired by the residual encoding approach described in \S\ref{sec:review}, we consider a block analogous to the residual computation of \eqref{eq:db residual} and \eqref{eq:query residual}. The approach consists of building a vector (denoting \relu\ as $\sigma$)
\begin{equation}
  \bm R_2 \sigma(\bm R_1 \bmt b'),
\end{equation}
that is analogous to the reconstruction $\bm d_n$ of $\bm x$ obtained from the encoding $n$ following the IVF stage (\cf \eqref{eq:bin assignment} and discussion thereof), and subtracts it from a linear mapping of $\bm x$:
\begin{equation}
  \bm r = \bm Q \bm x - \bm R_2 \sigma(\bm R_1 \bmt b').
\end{equation}
Besides the analogy to indexing pipelines, one other motivation for the above approach is to provide information to the subsequent feature encoding from the IVF bin selection stage (\ie $\bmt b'$) as well as the original feature $\bm x$. For completeness, as illustrated in \figref{fig:block diagram}, we also consider architectures that override this residual encoding block, setting $\bm r = \bm x$ directly.

The final stage 
consists of an $M$-block \subic\ encoder operating on $\bm r$ and producing test-time encodings $\bm b \in \ms K_K^M$, and training-time relaxed encoding $\bmt b \in \Delta_K^M$. Note that, unlike the residual approach described in \sxnref{sec:review}, our approach does not incurr the extra overhead required to compute LUTs using \eqref{eq:zluts}.

\paragraph{Searching} Given a query image $\bm I^*$, it is first fed to the pipeline in \figref{fig:block diagram} to obtain \textit{(i)} the activation coefficients $\bm z'^*$ at the output of the $\bm W_1$ layer and \textit{(ii)} the activation coefficients $\bm z^*$ at the output of the $\bm W_2$ layer. The IVF bins are then ranked 
as per \eqref{eq:sim sorting} and all database images 
$\big\{\bm I_i,~i\in \bigcup_{k=1}^B \ms B_{n_k}\big\}$ in the $B$ most pertinent bins are sorted, based on their encoding $\bm b_i$, 
according to their score
\begin{equation}
  \bm z'^{*\Tr} \bm b_i .
\end{equation}

\paragraph{Training} We assume we are given a training set $\{(\bm I^{(i)}, y^{(i)})\}_i$ organized into $C$ classes, where label $y^{(i)} \in (1,\ldots,C)$ specifies the class of the $i$-th image. 
Various works on learning for retrieval have explored the benefit of using ranking losses like the triplet loss and the pair-wise loss 
as opposed to the cross-entropy loss succesfully used in classification tasks \cite{gordo2016deep,xia2014supervised,zhang2015bit,zhao2015deep,lai2015simultaneous,lin2015deep,liu2016deep,do2016learning,Cagdas2015}. Empirically, we have found that the cross-entropy loss yields good results in the retrieval task, and we adopt it in this work.

Given an image belonging to class $c$ and a vector $\bm p \in \Delta_C$ that is an estimate of class membership probabilities, the cross-entropy loss is given by (the scaling is for convenience of hyper-parameter cross-validation)
\begin{equation}
  \ellT{\bm p, c} = -\frac{1}{\log_2{C}}\log_2 \bm p[c].
\end{equation}

Accordingly, we train our network by enforcing that the relaxed block-structured codes $\bmt b'$ and $\bmt b$ are good feature vectors that can be used to predict class membership. We do so by feeding each vector to a soft-max classification layer (layers $\bm C_1$ and $\bm C_2$ in \figref{fig:block diagram}, respectively), thus producing estimates of class membership $\bm s'$ and $\bm s$ in $\Delta_C$ (\cf \figref{fig:block diagram}) from which we derive two possible task-related losses. Letting $\ms T$ denote a batch specified as a set of training-pair indices, these two losses are 
%
%
\begin{gather}
  L_{1,\alpha}  = \frac{1}{|\ms T|} \sum_{i \in \ms T} \Big [  \alpha \ell\big(\bm s'^{(i)}, y^{(i)}\big) + \ell\big(\bm s^{(i)}, y^{(i)}\big) \Big] \\
  \textrm{ and } L_2  = \frac{1}{|\ms T|} \sum_{i \in \ms T}   \ell\big(\bm s'^{(i)} + \bm s^{(i)}, y^{(i)}\big) ,
\end{gather}
where the scalar $\alpha \in \{0,1\}$ is a selector variable.
In order to enforce the proximity of the $\bmt b'$ and $\bmt b$ to $\ms K_N$ and $\ms K_K^M$, respectively, we further employ the loss
\def\omegax{\Omega_{\ms H}}
\begin{multline}
  \omegax = \ellSx{\mu_1}{\gamma_1}{\{ \bmt b'^{(i)}  \}_{i \in \ms T}} +  \ellSx{\mu_2}{\gamma_2}{ \{ \bmt b^{(i)} \}_{i \in \ms T}},
  \label{eq:ent_losses_2}
\end{multline}
which depends on the four hyper-parameters $\ms H = \{\mu_1, \gamma_1, \mu_2, \gamma_2\}$ (we disuss heuristics for their selection if \sxnref{sec:exp}).

Accordingly, the general learning objective for our system is
\begin{equation}\label{eq:F*}
  F_* = L_* + \omegax,
\end{equation}
and we consider three variants thereof:
\begin{description}
\item
(\subicI) a non-residual variant  with objective $F_{1,1}$ corresponding to independently training the bin selection block and the feature encoder;

\item
(\subicR) a residual variant with objective $F_{1,0}$ where the bin selection block is pre-trained and held fixed during learning; and
  
\item
  (\subicJ) a non-residual variant with objective $F_2$.
\end{description}

\section{Experiments}\label{sec:exp}
\begin{table*}[h]
	\small
	\begin{center}
	\begin{tabu}{rcccccc} 
		Method & \oxf & \oxf* & \paris & \holidays & \oxff & \parisf \\ [-1pt] \tabucline[1pt]{1-7}
		DIR \cite{gordo2016deep} & 84.94 & 84.09 & 93.58 & 90.32 & 83.52 & 89.10\\ \hline
		PQ \cite{jegou2011product}& 46.57 & 39.45 & 57.57 & 48.23 & 38.73 & 42.23  \\ 
		\subic\ \cite{jain2017subic} & \textbf{53.25} & \textbf{46.06} & \textbf{71.28} & \textbf{60.52} & \textbf{46.88} & \textbf{58.27} \\ [-1pt] \tabucline[1pt]{1-7}
	\end{tabu}
	\end{center}
	\caption{\textbf{Instance retrieval with encoded features.} Performance (mAP) comparison using $64$-bit codes, first row shows reference results with original uncompressed features. When bounding box information is used for \oxf~dataset, the performance degrades for both PQ and \subic, shown in column \oxf*, as both are trained on full images.}\label{tab:retrieval}
\end{table*}

\begin{figure*}
	\centering
	\begin{tikzpicture}
          \pgfplotsset{
            with ylabel/.style={ylabel=mAP},
            with xlabel/.style={xlabel=$T$},
           with legend/.style={
             legend style={at={(0.5,-.5)}, legend columns=-2, anchor=north}, /tikz/every even column/.append style={column sep=1em} }
         }         
          \begin{groupplot}[
            group style={group size= 3 by 3, xlabels at=edge bottom, ylabels at=edge left, horizontal sep=20pt, vertical sep=30pt},
            height=4cm,width=6.5cm,xmode=log,
            cycle list name=mycyclelistA,
            title style={yshift=-.5em},
            ]
	\nextgroupplot[title=\oxf, with ylabel]
	\addplot table [x index=1, y index=2]{data/ivf-oxford5k.txt};	
	\addplot table [x index=3, y index=4]{data/ivf-oxford5k.txt};        
	\addplot table [x index=5, y index=6]{data/ivf-oxford5k.txt};        
	\addplot table [x index=7, y index=8]{data/ivf-oxford5k.txt};
	\addplot table [x index=9, y index=10]{data/ivf-oxford5k.txt};        
	\addplot table [x index=11, y index=12]{data/ivf-oxford5k.txt};        
	\nextgroupplot[title=\oxff]
	\addplot table [x index=1, y index=2]{data/ivf-oxford105k.txt};
	\addplot table [x index=3, y index=4]{data/ivf-oxford105k.txt};
	\addplot table [x index=5, y index=6]{data/ivf-oxford105k.txt};
	\addplot table [x index=7, y index=8]{data/ivf-oxford105k.txt};    
	\addplot table [x index=9, y index=10]{data/ivf-oxford105k.txt};
	\addplot table [x index=11, y index=12]{data/ivf-oxford105k.txt};
	\nextgroupplot[title=Oxford1M]
	\addplot table [x index=1, y index=2]{data/flickr1m-oxford.txt};
	\addplot table [x index=3, y index=4]{data/flickr1m-oxford.txt};
	\addplot table [x index=5, y index=6]{data/flickr1m-oxford.txt};
	\addplot table [x index=7, y index=8]{data/flickr1m-oxford.txt};    
	\addplot table [x index=9, y index=10]{data/flickr1m-oxford.txt};
	\addplot table [x index=11, y index=12]{data/flickr1m-oxford.txt};
	\nextgroupplot[title=\paris, with ylabel]
	\addplot table [x index=1, y index=2]{data/ivf-paris6k.txt};
	\addplot table [x index=3, y index=4]{data/ivf-paris6k.txt};
	\addplot table [x index=5, y index=6]{data/ivf-paris6k.txt};
	\addplot table [x index=7, y index=8]{data/ivf-paris6k.txt};   
	\addplot table [x index=9, y index=10]{data/ivf-paris6k.txt};
	\addplot table [x index=11, y index=12]{data/ivf-paris6k.txt};
	\nextgroupplot[title=\parisf]
	\addplot table [x index=1, y index=2]{data/ivf-paris106k.txt};
	\addplot table [x index=3, y index=4]{data/ivf-paris106k.txt};
	\addplot table [x index=5, y index=6]{data/ivf-paris106k.txt};
	\addplot table [x index=7, y index=8]{data/ivf-paris106k.txt};   
	\addplot table [x index=9, y index=10]{data/ivf-paris106k.txt};
	\addplot table [x index=11, y index=12]{data/ivf-paris106k.txt};
	\nextgroupplot[title=Paris1M]
	\addplot table [x index=1, y index=2]{data/flickr1m-paris.txt};
	\addplot table [x index=3, y index=4]{data/flickr1m-paris.txt};
	\addplot table [x index=5, y index=6]{data/flickr1m-paris.txt};
	\addplot table [x index=7, y index=8]{data/flickr1m-paris.txt};   
	\addplot table [x index=9, y index=10]{data/flickr1m-paris.txt};
	\addplot table [x index=11, y index=12]{data/flickr1m-paris.txt};
	\nextgroupplot[title=Holidays, with ylabel, with xlabel]
	\addplot table [x index=1, y index=2]{data/ivf-holidays.txt};
	\addplot table [x index=3, y index=4]{data/ivf-holidays.txt};
	\addplot table [x index=5, y index=6]{data/ivf-holidays.txt};
	\addplot table [x index=7, y index=8]{data/ivf-holidays.txt};
	\addplot table [x index=9, y index=10]{data/ivf-holidays.txt};
	\addplot table [x index=11, y index=12]{data/ivf-holidays.txt};
	\nextgroupplot[title=Holidays101K, with xlabel, with legend] 
	\addplot table [x index=1, y index=2]{data/ivf-holidays101k.txt};
	\addplot table [x index=3, y index=4]{data/ivf-holidays101k.txt};
	\addplot table [x index=5, y index=6]{data/ivf-holidays101k.txt};
	\addplot table [x index=7, y index=8]{data/ivf-holidays101k.txt};    
	\addplot table [x index=9, y index=10]{data/ivf-holidays101k.txt};
	\addplot table [x index=11, y index=12]{data/ivf-holidays101k.txt};
        \legend{\IVFPQ, \IMIPQ, DSH-\subic, \subicI, \subicJ, \subicR}
	\nextgroupplot[title=Holidays1M, with xlabel]
	\addplot table [x index=1, y index=2]{data/flickr1m-holidays.txt};
	\addplot table [x index=3, y index=4]{data/flickr1m-holidays.txt};
	\addplot table [x index=5, y index=6]{data/flickr1m-holidays.txt};
	\addplot table [x index=7, y index=8]{data/flickr1m-holidays.txt};    
	\addplot table [x index=9, y index=10]{data/flickr1m-holidays.txt};
	\addplot table [x index=11, y index=12]{data/flickr1m-holidays.txt};
	\end{groupplot}
	\end{tikzpicture}
	\caption{\textbf{Large-scale image retrieval with complete pipelines} Plots of mAP vs. (average) shortlist size $T$. For all methods except \IMIPQ, the $n$-th plotted point is obtained from all images in the first $B=2^n$ bins. For \IMIPQ, the mAP is computed on the first $T$ responses.} \label{fig:ivf}
\end{figure*}
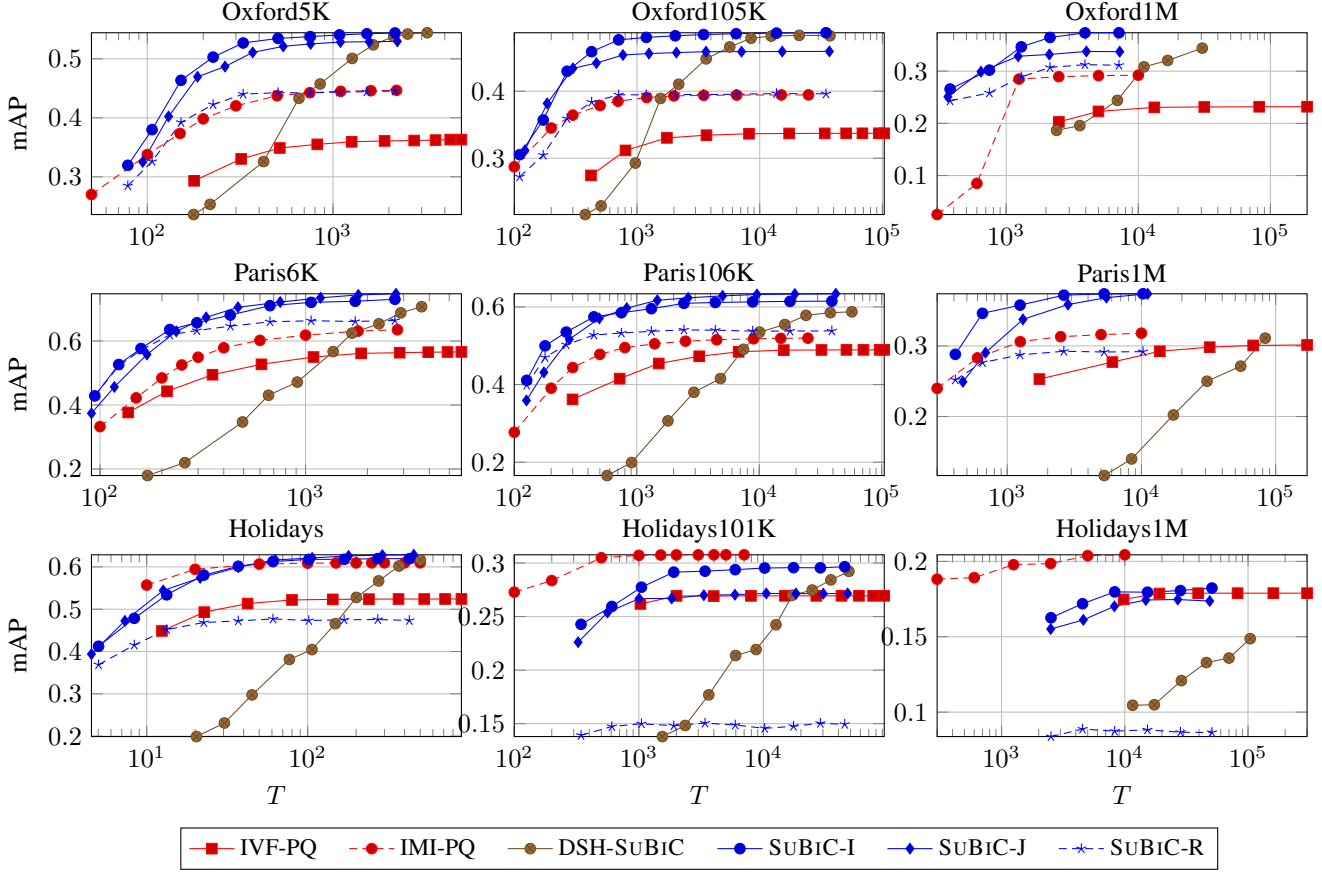

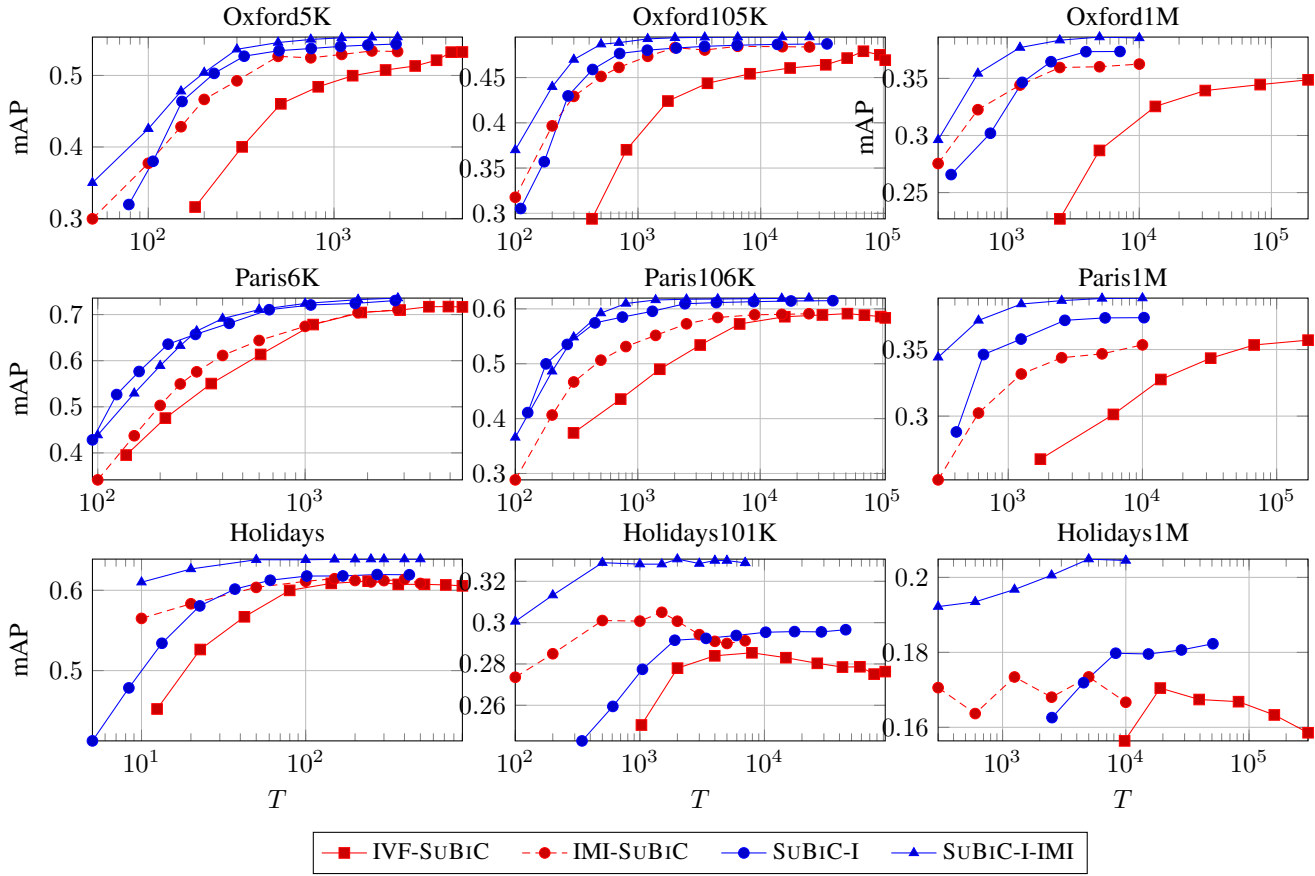
\begin{figure*}
	\centering
	\begin{tikzpicture}
	\pgfplotsset{
		with ylabel/.style={ylabel=mAP},
		with xlabel/.style={xlabel=$T$},
		with legend/.style={
			legend style={at={(0.5,-.5)}, legend columns=-2, anchor=north}, /tikz/every even column/.append style={column sep=1em} }}
                    \begin{groupplot}[group style={group size= 3 by 3, xlabels at=edge bottom, ylabels at=edge left, horizontal sep=20pt, vertical sep=30pt},
                      height=4cm,width=6.5cm, xmode=log,title style={yshift=-.5em},
                      cycle list name=mycyclelistB ]
	\nextgroupplot[title=\oxf, with ylabel, legend style = { column sep = 2pt, legend columns = -1, legend to name = grouplegend,}]
	\addplot table [x index=13, y index=14]{data/ivf-oxford5k.txt}; 
	\addplot table [x index=15, y index=16]{data/ivf-oxford5k.txt};	
	\addplot table [x index=7, y index=8]{data/ivf-oxford5k.txt};        
	\addplot table [x index=0, y index=1]{data/subic-I-imi.txt};
	\nextgroupplot[title=\oxff]
	\addplot table [x index=13, y index=14]{data/ivf-oxford105k.txt}; 
	\addplot table [x index=15, y index=16]{data/ivf-oxford105k.txt};	
	\addplot table [x index=7, y index=8]{data/ivf-oxford105k.txt};        
	\addplot table [x index=0, y index=1]{data/subic-I-imi-100k.txt};
	\nextgroupplot[title=Oxford1M, with ylabel]
	\addplot table [x index=13, y index=14]{data/flickr1m-oxford.txt}; 
	\addplot table [x index=15, y index=16]{data/flickr1m-oxford.txt};	
	\addplot table [x index=7, y index=8]{data/flickr1m-oxford.txt};        
	\addplot table [x index=0, y index=1]{data/subic-I-imi-1M.txt};
	\nextgroupplot[title=\paris, with ylabel]
	\addplot table [x index=13, y index=14]{data/ivf-paris6k.txt};
	\addplot table [x index=15, y index=16]{data/ivf-paris6k.txt};
	\addplot table [x index=7, y index=8]{data/ivf-paris6k.txt};
	\addplot table [x index=2, y index=3]{data/subic-I-imi.txt};
	\nextgroupplot[title=\parisf]
	\addplot table [x index=13, y index=14]{data/ivf-paris106k.txt};
	\addplot table [x index=15, y index=16]{data/ivf-paris106k.txt};
	\addplot table [x index=7, y index=8]{data/ivf-paris106k.txt};
	\addplot table [x index=2, y index=3]{data/subic-I-imi-100k.txt};
	\nextgroupplot[title=Paris1M]
	\addplot table [x index=13, y index=14]{data/flickr1m-paris.txt};
	\addplot table [x index=15, y index=16]{data/flickr1m-paris.txt};
	\addplot table [x index=7, y index=8]{data/flickr1m-paris.txt};
	\addplot table [x index=2, y index=3]{data/subic-I-imi-1M.txt};
	\nextgroupplot[title=Holidays, with ylabel, with xlabel]
	\addplot table [x index=13, y index=14]{data/ivf-holidays.txt};
	\addplot table [x index=15, y index=16]{data/ivf-holidays.txt};
	\addplot table [x index=7, y index=8]{data/ivf-holidays.txt};
	\addplot table [x index=4, y index=5]{data/subic-I-imi.txt};
	\nextgroupplot[title=Holidays101K, with xlabel, with legend]
	\addplot table [x index=13, y index=14]{data/ivf-holidays101k.txt};
	\addplot table [x index=15, y index=16]{data/ivf-holidays101k.txt};
	\addplot table [x index=7, y index=8]{data/ivf-holidays101k.txt};
	\addplot table [x index=4, y index=5]{data/subic-I-imi-100k.txt};
			\legend{IVF-\subic, IMI-\subic, \subicI, \subicI-IMI}
	\nextgroupplot[title=Holidays1M, with xlabel]
	\addplot table [x index=13, y index=14]{data/flickr1m-holidays.txt};
	\addplot table [x index=15, y index=16]{data/flickr1m-holidays.txt};
	\addplot table [x index=7, y index=8]{data/flickr1m-holidays.txt};
	\addplot table [x index=4, y index=5]{data/subic-I-imi-1M.txt}; 
	\end{groupplot}
	\end{tikzpicture}
	\caption{\textbf{IMI variant of our approach and comparison for fixed encoder} Comparison of an IMI variant of our method to the baselines, when using the same (non-residual) feature encoder. Note the substantial relative improvements of \subicI-IMI.}\label{fig:subic-i}
\end{figure*}

\paragraph{Datasets} For large-scale image retrieval, we use three publicly available datasets to evaluate our approach: \oxf~\cite{philbin2007object}\footnote{\url{www.robots.ox.ac.uk/~vgg/data/oxbuildings/}}, \paris~\cite{philbin2008lost}\footnote{\url{www.robots.ox.ac.uk/~vgg/data/parisbuildings/}} and \holidays~\cite{jegou2008hamming}\footnote{\url{lear.inrialpes.fr/~jegou/data.php}}. For large-scale experiments, we add 100K and 1M images from Flickr (\flickr\ and \flickr1M\ respectively) as a noise set.
For \oxf, bounding box information is not used. For Holidays, images are used without correcting orientation.

For training, we use the \landm-full subset of the \landm~dataset \cite{babenko2014neural}\footnote{\url{sites.skoltech.ru/compvision/projects/neuralcodes/}}, as in \cite{gordo2016deep}. We could only get 125,610 images for the full set due to broken URLs. In all our experiments and for all approaches we use \landm-full as the training set.

For completeness, we also carry out \emph{category retrieval} \cite{jain2017subic} test using the \voc \footnote{\url{http://host.robots.ox.ac.uk/pascal/VOC/}} and \caltech \footnote{\url{http://www.vision.caltech.edu/Image_Datasets/Caltech101/}} datasets. For this test, our method is trained on ImageNet. 

\paragraph{Base features} The base features $\mv x$ are obtained from the
ResNet version of the network proposed in \cite{gordo2016deep}. This network
extends the ResNet-101 architecture with region of interest pooling, fully connected
layers, and $\ell_2$-normalizations to mimic the pipeline used
for instance retrieval. Their method enjoys state-of-the-art 
performance for instance retrieval, motivating its usage as
base CNN for this task.

\paragraph{Hyper-parameter selection} For all three variants of our approach (\subicI, \subicJ, and \subicR), we use $N=4096$ bins, and a \subic-$(8,256)$ encoder having $M=8$ blocks of $K=256$ block size (corresponding to $8$ bytes per encoded feature). These parameters correspond to commonly used values for indexing systems. To select the four hyper-parameters $\ms H = \{\mu_1, \gamma_1, \mu_2, \gamma_2\}$ (\cf \eqref{eq:F*}) we first cross-validate just the bin selection block to choose $\mu_1=5.0$ and $\gamma_1=6.0$. With these values fixed, we then cross-validate the encoder block to obtain $\mu_2=0.6$ and $\gamma_2=0.9$. We use the same values for all three variants of our system.
%

\paragraph{Evaluation of feature encoder} First of all, we evaluate how \subic~encoding performs on all the test datasets compared to the PQ unsupervised vector quantizer. We use $M=8$ and $K=256$ setups for both codes. 
\subic~is trained for 500K batches of 200 training images, with $\mu=0.6$ and $\gamma=0.9$. 
The results reported in Table \ref{tab:retrieval} show that, as expected, \subic~outperforms PQ, justifying its selection as a feature
encoder in our system. For reference, the first row in the
table gives the performance with uncompressed features. While high, each base
feature vector has a storage footprint of
8 Kilo bytes (assuming 4-byte floating points). \subic\ and
PQ, on the other hand, require only 8 bytes of storage per
feature ($1000\times$ less).


\paragraph{Baseline indexing systems} We compare all three variants
of our proposed indexing system against two existing
baselines, as well as a straightforward attempt to use
deep hashing as an IVF system:

\begin{description}
\item (IVF-PQ) This approach uses an inverted file with $N =
4096$ bins followed by a residual PQ encoder with $M = 8$
blocks and constituent codebooks of size $K = 256$ (\cf~\eqref{eq:constituent}), 
resulting in an 8-byte feature size. The search employs
asymmetric distance computation. During retrieval,
the top $B = 2^n$, lists are retrieved, and, for each $n =1,2,\ldots $
the average mAP and average aggregate bin size $T$ are plotted.

\item(IMI-PQ) The Inverted Multi-Index (IMI) \cite{babenko12inverted} extends the
standard IVF by substituting a product quantizer with $M =
2$ and $K = 4096$ in place of the vector quantizer. The resulting
IVF has more than 16 million bins, meaning that,
for practical testing sets (containing close to 1 million images),
most of the bins are empty. Hence, when employing
IMI, we select shortlist sizes $T$ for which to compute average
mAP to create our plots. Note that, given the small size
of the IMI bins, the computation of the look-up tables $\bm z_n$
(\cf \eqref{eq:zluts}) represents a higher cost per-image for IMI
than for IVF. Furthermore, the fragmented memory reads
required can have a large impact on speed relative to the
contiguous reads implicit in the larger IVF bins.



\item(DSH-\subic) In order to explore possible approaches to
include supervision in the IVF stage of an indexing system,
we further considered using the DSH deep hash code \cite{liu2016deep} as
a bin selector, carefully selecting the regularization parameter
to be 0.03 by means of cross-validation.
We train this network to produce 12-bit image representations
corresponding to $N = 4096$ IVF bins, where each bin
has an associated hash code. Images are indexed using their
DSH hash, and at query time, the Hamming distance between
the query's 12-bit code and each bin's code is used to rank
the lists. For the encoder part, we used \subic\ with $M = 8$
and $K = 256$, the same used in Tab. \ref{tab:retrieval}.

\end{description}

\paragraph{Large-scale indexing} Fig \ref{fig:ivf} shows the mAP performance
versus average number of retrieved images $T$ for
all three variants as well as the baselines described above.
Note that the number of retrieved images is a measure of
complexity, as for IVF, the time complexity of the system
 is dominated by the approximate distance computations in \eqref{eq:lut}. 
For IMI, on the other hand, there is a non-negligible overhead on top of the approximate distance computation related to the large
number of bins, as discussed above.


We present results for three datastets (\oxf, \paris, \holidays), on three different database scales (the original dataset, and when also including noise datasets of $100K$ and $1M$ images). Note that on \oxf\ and \paris, both \subicI\ and \subicJ\ enjoy large advantages relative to all three baselines -- at $T=300$, the relative advantage of \subicI\ over the \IMIPQ\ is 19\% at least. \subicR\ likewise enjoys an advangate on the \paris\ dataset, and performs comparably to the baselines on \oxf. 

On \holidays\, \subicI\ outperforms \IVFPQ\ by a large margin ($18\%$ relative), but not outperform \IMIPQ. As discussed above, this comparison does not reflect the overhead implicit in an IMI index. To illustrate this overhead, we note that, when 1M images are indexed, the average (non-empty) bin size for IMI is 18.3, meaning that approximately 54.64 memory accesses and look-up table constructions need to be carried out for each IMI query per 1K images. This compares to an average bin size of 244.14 for IVF, and accordingly 4.1 contiguous memory reads and look-up table constructions. Note, on the other hand, that \subicI\ readily outperforms \IVFPQ\ in all \holidays\ experiments.

Concerning the poor performance of \subicR\ on \holidays, we believe this is due to poor generalization ability of the system because of the three extra fully-connected layers.

\paragraph{IMI extension} Given the high performance of IMI for the \holidays\ experiments in \figref{fig:ivf}, we further consider an IMI variant of our \subicI\ architecture. To implement this approach, we learn a \subic-$(2,4096)$ encoder (with $\mu=4$ and $\gamma=5$). Letting $\bm z_m'$ denote the $m$-th block of $\bm z'$, the $(k,l) \in (1,\ldots, 4096)^2$  bins of \subic-IMI are sorted based on the score $\bm z'_1[k] + \bm z'_2[l]$. For fairness of comparison, we use the same \subic$-(8,256)$ feature encoder for all methods including the baselines, which are IVF and IMI with unsupervised codebooks (all methods are non-residual). The results, plotted in \figref{fig:subic-i}, establish that, for the same number of bins, our method can readily outperform the baseline IMI (and IVF) methods. Furthermore, given that we use the best performing feature encoder (\subic) for all methods, this experiment also establishes that the \subic\ based binning system that we propose outperforms the unsupervised IVF and IMI baselines.

\begin{figure}
  \centering
   \begin{tikzpicture}
             \pgfplotsset{
   	with ylabel/.style={ylabel=mAP},
   	with xlabel/.style={xlabel=$T$},
      }
      \begin{groupplot}[group style={group size= 2 by 1, xlabels at=edge bottom, ylabels at=edge left, horizontal sep=20pt, vertical sep=0pt},
        height=4cm,width=.57\columnwidth,xmode=log, title style={yshift=-.5em},
        cycle list name=mycyclelistC
        ]
   	     \nextgroupplot[ymin=0.2, title=\caltech, with ylabel, with xlabel, legend pos=south east, y label style={inner sep=0, at={(axis description cs:0.15,.5)},anchor=south}, legend style={draw=none, fill=none, at={(1,0)}, anchor=south east}]
   	     \addplot table [x index=1, y index=2]{data/caltech.txt};
   	     \addlegendentry{IVFPQ}
   	     \addplot table [x index=3, y index=4]{data/caltech.txt};
   	     \addlegendentry{\subic-I}
   	     \addplot table [x index=5, y index=6]{data/caltech.txt};
   	     \addlegendentry{\subic-R}
   	     \nextgroupplot[ymax=0.4, title=\voc, with xlabel]
   	     \addplot table [x index=1, y index=2]{data/pascal.txt};
   	     \addplot table [x index=3, y index=4]{data/pascal.txt};
   	     \addplot table [x index=5, y index=6]{data/pascal.txt};
   	     \end{groupplot}
   	\end{tikzpicture}
   	\caption{\textbf{ Category retrieval} Comparing \subicI\ and \subicR\ to \IVFPQ\ on the category retrieval task. joint-residual to non-joint \subic~and IVFPQ for category retrieval on Pascal and Caltech101. All methods are trained on \textit{VGG-M-128} features of ImageNet images.}\label{fig:cat}
\end{figure}
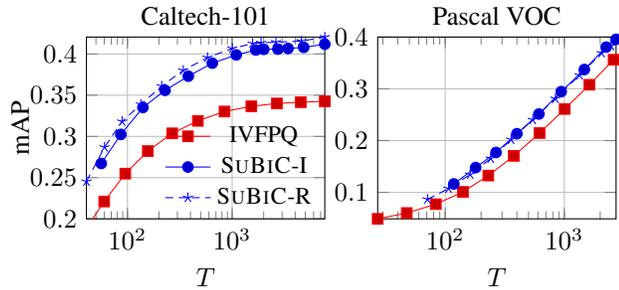
\paragraph{Category retrieval} For completeness, we also carry out experiments in the category retrieval task which has been the main focus of recent deep hashing methods \cite{xia2014supervised,zhang2015bit,zhao2015deep,lai2015simultaneous,lin2015deep,liu2016deep,do2016learning}. In this task, a given query image is used to rank all database images, with a correct match occuring for database images of the same class as the query image. For category retrieval experiments, we use VGG-M-128 base features \cite{simonyan2014very}, which have established good performance for classification tasks, and a \subic-$(1,8192)$ for bin selection. We use the ImageNet training set (1M+ images) to train, and the test (training) subsets of \voc\ and \caltech\ as a query (respectively, database) set. We present results for this task in \figref{fig:cat}. Note that, unlike the \holidays\ experiments in \figref{fig:ivf}, \subicR\ performs best on \caltech\ and equally well to \subicI\ on \voc, a consequence of the greater size and diversity of the ImageNet datasets relative to the \landm\ dataset.


\section{Conclusion}\label{sec:conclusion}

We present a full image indexing pipeline that exploits supervised deep learning methods to build an inverted file as well as a compact feature encoder. Previous methods have either employed unsupervised inverted file mechanisms, or employed supervision only to derive feature encoders. We establish experimentally that our method achieves state of the art results in large scale image retrieval.

{\small
\bibliographystyle{ieee}
\bibliography{egbib}
}

\end{document}